\documentclass[runningheads]{llncs}
\usepackage[T1]{fontenc}
\usepackage{graphicx,verbatim}
\usepackage{booktabs}
\usepackage{multirow}
\usepackage{xcolor}
\usepackage{float}      
\usepackage{caption}
\begin{document}
\title{RADAR: A Multimodal Benchmark for 3D Image-Based Radiology Report Review}
\titlerunning{RADAR: Multimodal Benchmark for Report Review}

\author{Zhaoyi Sun\inst{1}
\and
Minal Jagtiani\inst{1}
\and
Wen-wai Yim\inst{2}
\and
Fei Xia\inst{1}
\and
Martin Gunn\inst{1}
\and\\
Meliha Yetisgen\inst{1}\textsuperscript{\textdagger}
\and
Asma Ben Abacha\inst{2}\textsuperscript{\textdagger}
}
\authorrunning{Z. Sun et al.}
\institute{University of Washington, Seattle, WA 98195, USA\\ 
\email{\{zhaoyis,minalj,fxia,marting,melihay\}@uw.edu}\\
\and
Microsoft Health AI, Redmond, WA 98052, USA\\
\email{\{yimwenwai,abenabacha\}@microsoft.com}}





  
\maketitle

\begingroup
\renewcommand\thefootnote{\textdagger}
\footnotetext{These authors contributed equally as senior authors.}
\endgroup

\setcounter{footnote}{0}

\begin{abstract}

Radiology reports for the same patient examination may contain clinically meaningful discrepancies arising from interpretation differences, reporting variability, or evolving assessments. Systematic analysis of such discrepancies is important for quality assurance, clinical decision support, and multimodal model development, yet remains limited by the lack of standardized benchmarks. We present RADAR, a multimodal benchmark for radiology report discrepancy analysis that pairs 3D medical images with a preliminary report and corresponding candidate edits for the same study. The dataset reflects a standard clinical workflow in which trainee radiologists author preliminary reports that are subsequently reviewed and revised by attending radiologists. RADAR defines a structured discrepancy assessment task requiring models to evaluate proposed edits by determining image-level agreement, assessing clinical severity, and classifying edit type (correction, addition, or clarification). In contrast to prior work emphasizing binary error detection or comparison against fully independent reference reports, RADAR targets fine-grained clinical reasoning and image–text alignment at the report review stage. The benchmark consists of expert-annotated abdominal CT examinations and is accompanied by standardized evaluation protocols to support systematic comparison of multimodal models. RADAR provides a clinically grounded testbed for evaluating multimodal systems as reviewers of radiology report edits. 

\keywords{Radiology report evaluation  \and  Multimodal benchmark \and VLM.}

\end{abstract}

\section{Introduction}
Radiology report discrepancies are a recognized concern for patient safety and quality assurance, occurring in approximately 3–5\% of routine interpretations, with higher rates reported in emergency department (ED) imaging~\cite{Brady2017-hm}. In ED settings, particularly within academic centers where preliminary resident reports are later reviewed by attendings, even small discrepancy rates can have disproportionate downstream impact, as clinical decisions may be made before the final interpretation is available~\cite{Friedman2013-jw,Ruchman2007-fd}. Large retrospective studies have reported approximately 2.6\% of emergency imaging interpretations contained discrepancies requiring direct physician notification due to potential impact on patient care, with contrast-enhanced abdominal CT among the most commonly involved examinations~\cite{Ruchman2007-fd}. Although many discrepancies do not alter management, a clinically meaningful subset is associated with negative or significant effects on patient care, underscoring the need for systematic discrepancy analysis.

Despite rapid progress in medical AI, existing datasets and benchmarks for radiology discrepancy or error analysis remain limited, particularly for volumetric CT. Many prior approaches consider errors as text-only phenomena or rely on synthetically introduced errors (e.g., word insertion, deletion, or substitution), which may not faithfully represent clinically meaningful, image-evidenced interpretive discrepancies and are often not detectable without true image understanding~\cite{Sun2025-nf,Zech2019-ix}. Meanwhile, large-scale 3D CT vision–language datasets and benchmarks developed for report generation or visual question answering (VQA) are not designed around the real-world preliminary-to-final reporting workflow and do not directly evaluate whether a proposed edit is supported by imaging evidence or resolves a discrepancy. To the best of our knowledge, this work presents the first multimodal benchmark\footnote{The benchmark will be made available upon request and completion of a data usage agreement, in compliance with institutional and hospital data governance policies.} specifically designed for image-grounded radiology report discrepancy analysis that conditions on a CT image, a preliminary report, and a candidate suggested edit. Our contributions are as follows:

\begin{itemize}
    \item A multimodal benchmark derived from real trainee-to-attending report revisions, rather than synthetically introduced errors, enabling evaluation of whether candidate edits are supported by imaging evidence.
    
    \item A fine-grained evaluation framework that jointly assesses agreement, clinical severity, and discrepancy type.
    
    \item An empirical study of several vision–language foundation models under multiple volumetric input settings, establishing a baseline for future methods on image-grounded discrepancy analysis.
\end{itemize}


\section{Related Work}
\subsection{Quality Assurance and Error Detection}
Radiology report discrepancies have been widely studied in the context of quality assurance, education, and clinical risk mitigation. Lyo et al.~\cite{Lyo2025-sd} analyzed preliminary-to-final report revisions and showed that generative models can convert real-world edits into structured educational feedback, highlighting the value of discrepancy analysis. Recent studies have extended this perspective to multimodal error detection by incorporating imaging data. Wu et al.~\cite{Wu2023-rb} evaluated multimodal LLMs on a benchmark constructed by injecting synthetic insertion, removal, and substitution errors into radiology reports paired with images; although fine-tuned models achieved clinician-level performance in binary detection, they struggled with fine-grained classification. MedErr-CT introduced a CT-specific VQA benchmark and evaluated several 3D multimodal models, reporting substantial performance gaps across error types and task levels~\cite{Kyung2025-ab}. However, these studies rely primarily on synthetic errors and constrained tasks, and do not explicitly evaluate whether report edits are image-supported.

A substantial body of work focuses on text-only error detection. Early NLP approaches modeled insertion, substitution, and deletion errors using sequence-to-sequence or classification architectures, showing that neural models can detect implausible report content~\cite{Zech2019-ix}. More recently, GPT-based and fine-tuned LLMs have been applied to identify factual inconsistencies and improve reporting accuracy~\cite{Gertz2024-rg,Salam2025-zh,Sun2025-nm}. Other efforts such as GREEN, which introduces structured error annotation and factuality-aware metrics~\cite{ostmeier-etal-2024-green}, and ReXErr, which synthesizes clinically meaningful errors~\cite{Rao2024-lr}, further standardize report evaluation. Although these text-centric approaches do not verify image evidence, they provide insights into discrepancy characterization and evaluation methodology.

\subsection{Vision-Language Models (VLMs) in 3D Radiology}

Vision-language modeling in 3D radiology can be grouped by input representation. Volumetric models (e.g., RadFM~\cite{Wu2025-fw}, CT-CLIP~\cite{Hamamci2025-xt}, Merlin~\cite{Blankemeier2024-du}, M3D-LaMed~\cite{Bai2024-co}) employ native 3D encoders trained with multimodal objectives~\cite{Wu2025-fw,Hamamci2025-xt,Blankemeier2024-du,Bai2024-co}. Slice-based approaches instead adapt 2D VLM backbones using slice pooling or selection strategies.

Recent medical multimodal and generalist models (e.g., Lingshu~\cite{LASA-Team2025-ak}, Hulu-Med~\cite{Jiang2025-lo}, Gemini-3-pro, GPT-5.2, Qwen3.5-plus) operate on sampled slices rather than volumetric tokens~\cite{LASA-Team2025-ak,Jiang2025-lo}. However, cross-paradigm comparisons remain limited. Many studies do not benchmark against strong slice-based baselines under matched settings, and reported gains often reflect data scale and supervision rather than architecture alone. The relative advantages of native 3D encoding versus slice-based prompting therefore remain unclear.



\subsection{Datasets for 3D Radiology Report Generation and VQA}
Progress in 3D CT vision–language modeling has been driven by the release of datasets pairing volumetric scans with reports. CT-RATE, comprising 25,692 non-contrast 3D chest CT scans with reports, has become a foundational benchmark for models such as CT-CLIP and CT-CHAT~\cite{Hamamci2025-xt}. RadGenome-Chest CT augments CT-RATE with organ-level segmentation masks, sentence-level grounding annotations, and grounded VQA pairs, enabling regionally grounded reasoning and fine-grained evaluation~\cite{zhang2024radgenome}. Despite these advances, publicly available 3D CT datasets remain limited in scale and diversity compared to 2D radiography benchmarks, constraining systematic evaluation of volumetric VLMs.

Alongside dataset development, several studies have explored 3D CT report generation and VQA. Li et al.~\cite{Li2025-ej} proposed a brain CT report generation framework and introduced FORTE, a feature-oriented evaluation scheme designed to better reflect diagnostic quality than text-overlap metrics. Other approaches incorporate explicit anatomical or spatial modeling, such as CT-GRAPH, which encodes anatomical hierarchies for report generation on CT-RATE~\cite{Kalisch2025-wv}, and 3D-CT-GPT, which frames report generation through VQA-style conditioning~\cite{Chen2024-vu}. In 3D CT VQA, CT-Agent adopts an agentic framework to address long-range spatial dependencies across slices~\cite{Mao2025-mn}, while M3 extends multimodal modeling to abdominal CT~\cite{Hosseini2025-ue}. Although these efforts focus on generation and question answering rather than discrepancy resolution, they highlight challenges in volumetric reasoning that motivate more targeted benchmarks.

\section{Task Definition}
\begin{figure}[!h]
    \centering
    \includegraphics[width=\linewidth]{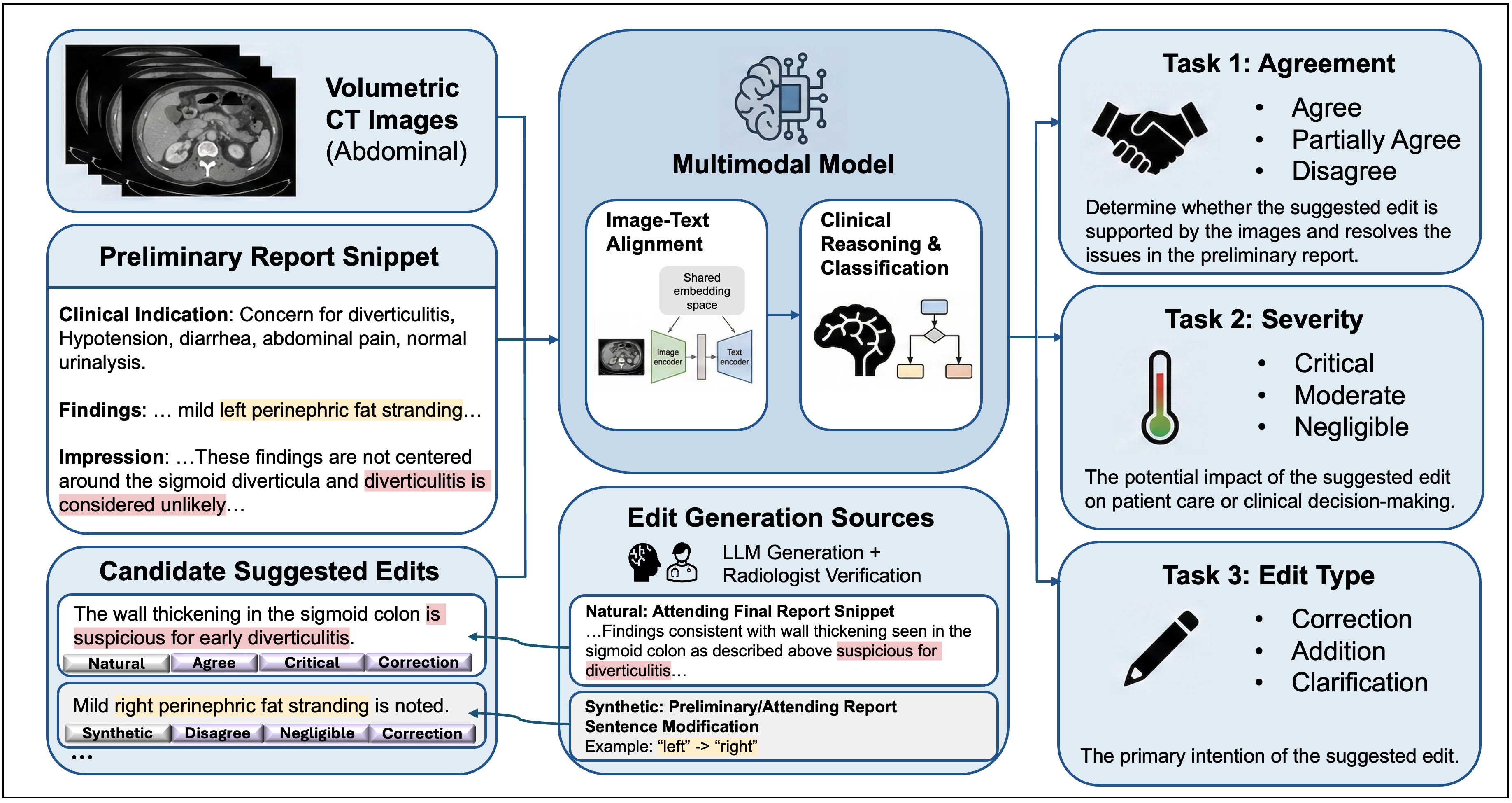}
    \caption{Overview of the RADAR workflow.}
    \label{fig:workflow}
\end{figure}
Given volumetric CT images, a preliminary report, and a candidate suggested edit referring to the same study, the task is to evaluate whether the suggested edit is supported by the imaging evidence and resolves a discrepancy in the preliminary report. Specifically, models are required to: (1) determine the level of image-grounded agreement of the suggested edit (agree, partially agree, or disagree), (2) assess the clinical severity of the underlying discrepancy (critical, moderate, or negligible), and (3) classify the edit type as a correction, addition, or clarification. The task is formulated to reflect real-time report review, where models evaluate candidate edits without access to the finalized attending report. Figure~\ref{fig:workflow} illustrates the RADAR workflow and task structure.

\begin{table*}[!t]
\centering
\caption{Dataset characteristics and annotation statistics. \textit{Natural} denotes discrepancies arising from preliminary-to-attending report revisions. \textit{Mixed} includes natural discrepancies and disagree edits introduced for evaluation balancing.}
\label{tab:dataset_stats}
\scriptsize
\setlength{\tabcolsep}{4pt}

\begin{minipage}[t]{0.48\textwidth}
\centering
\textbf{(a) Case-level Statistics}
\begin{tabular}{lr}
\toprule
\multicolumn{2}{l}{\textbf{Number of Cases}} \\
\midrule
Total CT examinations & 50 \\
Validation cases & 20 \\
Test cases & 30 \\
Daytime cases & 25 \\
Overnight cases & 25 \\
\midrule
\multicolumn{2}{l}{\textbf{Report Length (words), mean (SD)}} \\
\midrule
\textit{Preliminary} \\
Daytime & 220.48 (66.74) \\
Overnight & 88.00 (38.28) \\
\midrule
\multicolumn{2}{l}{\textbf{Imaging Statistics}} \\
\midrule
Image series & 319 \\
Slices & 45,838 \\
\bottomrule
\end{tabular}
\end{minipage}
\hfill
\begin{minipage}[t]{0.48\textwidth}
\centering
\textbf{(b) Edit-level Statistics}
\scriptsize
\setlength{\tabcolsep}{3pt}

\begin{tabular*}{\textwidth}{@{\extracolsep{\fill}}lcccc@{}}
\toprule
& \multicolumn{2}{c}{\textbf{Natural}} & \multicolumn{2}{c}{\textbf{Mixed}} \\ 
\cmidrule(lr){2-3} \cmidrule(lr){4-5}
\textbf{Count} & \textbf{Eval} & \textbf{Test} & \textbf{Eval} & \textbf{Test} \\
\midrule
\textit{Total candidate edits} & 102 & 153 & 184 & 274 \\
\midrule
\textit{Agreement} & & & & \\
Agree & 88 & 132 & 88 & 132 \\
Partially Agree & 11 & 16 & 11 & 16 \\
Disagree & 3 & 5 & 85 & 126 \\
\midrule
\textit{Severity} & & & & \\
Negligible & 67 & 117 & 94 & 141 \\
Moderate & 25 & 27 & 45 & 69 \\
Critical & 10 & 9 & 45 & 64 \\
\midrule
\textit{Edit Type} & & & & \\
Addition & 61 & 110 & 108 & 192 \\
Correction & 27 & 28 & 42 & 51 \\
Clarification & 14 & 15 & 34 & 31 \\
\bottomrule
\end{tabular*}
\raggedright
\end{minipage}
\end{table*}

\section{Dataset Creation}
\subsection{Case Selection and Data Sources}
We collected abdomen and pelvis CT examinations from the ED at Harborview Medical Center (HMC), UW Medicine. Each examination includes all CT image series (DICOM format) along with paired preliminary and attending final radiology reports. Preliminary reports are authored by residents or fellows and are available continuously (24/7), while attending radiologists subsequently issue final reports after reviewing both the imaging and the preliminary interpretation. This study was conducted under Institutional Review Board (IRB) approval.

The dataset includes both daytime and overnight ED cases, where the designation (daytime vs. overnight) refers to the time at which the preliminary report was authored. Daytime preliminary reports are typically more comprehensive and later refined by attendings, whereas overnight reports often follow a “quick-look” workflow in which residents focus on identifying acute findings requiring immediate attention, with attendings subsequently expanding or correcting the interpretation. Table \ref{tab:dataset_stats}(a) shows case-level statistics of RADAR. All data were de-identified. We treat the attending final report as the reference standard for identifying discrepancies relative to the preliminary report. 

\subsection{Preprocessing and Candidate Edit Generation}
Candidate suggested edits were derived by first identifying discrepancies between preliminary and final reports using GPT-4o, followed by manual review by an attending radiologist. The radiologist rejected incorrect suggestions, added missed discrepancies, and reformulated each verified discrepancy into a concise edit intended to resolve the difference in the preliminary report.

Edits referring to longitudinal trends or prior examinations were excluded, as they cannot be verified from a single CT study.

\subsection{Gold-Standard Annotation and Dataset Balancing}
Each candidate edit was annotated by a radiologist with 10 years of post-board-certification experience along three dimensions: \textit{agreement}, \textit{severity}, and \textit{edit type}. \textit{Agreement} indicates whether the edit is factually accurate and supported by the imaging evidence (agree), generally correct but minor or redundant (partially agree), or incorrect/unnecessary (disagree). \textit{Severity} reflects the potential clinical impact relative to the indication (critical, moderate, or negligible). \textit{Edit type} captures the primary intention of the edit (correction, addition, or clarification) regardless of its accuracy. Dataset statistics are summarized in Table \ref{tab:dataset_stats}.

We observed substantial class imbalance in the natural dataset (220 agree, 27 partially agree, 8 disagree). To support balanced evaluation, we introduced a set of synthetic disagree edits used solely to balance the evaluation split and assess model robustness to unsupported edits. These were generated using GPT-5.2 to produce plausible but incorrect modifications of otherwise valid report statements, guided by previously described error patterns (e.g., ReXErr~\cite{Rao2024-lr}) and the empirical edit type distribution of natural discrepancies. All synthetic edits were manually reviewed and annotated by a board-certified radiologist to confirm the correct label as disagree. As summarized in Table \ref{tab:dataset_stats}(b), \textit{Natural} refers to discrepancies arising from real preliminary-to-attending revisions, whereas \textit{Synthetic} denotes model-generated but expert-verified incorrect edits added for evaluation balance.

\section{Methods}
\subsection{Baseline Systems}
We evaluate several multimodal foundation models on RADAR. Prompts, preprocessing steps, and inference configurations are included in released codebase.

\begin{itemize}
    \item \textbf{Series Selection.} Because CT studies contain multiple series, we first select the most relevant series for each candidate edit using GPT-OSS-20B. The model receives structured DICOM metadata (e.g., anatomical region, acquisition timing) and the edit text, and outputs the best-matched series (e.g., COR BODY). Subsequent inputs are restricted to the selected series.
    
    \item \textbf{Gemini-2.5-Pro and Gemini-3-Pro.} Multimodal models developed by Google. Evaluated with four visual input settings: 50, 20, or 10 uniformly sampled slices, and a video constructed from all slices.
    
    \item \textbf{Qwen3.5-plus.} Multimodal model developed by Alibaba. Evaluated under the same four slice/video configurations for direct comparison.
\end{itemize}

\subsection{Evaluation Metrics}

We report metrics under two evaluation settings that differ only in the test subset. Accuracy is reported for agreement, severity, and edit type. We additionally compute a \textit{Composite Score} to jointly assess agreement and fine-grained reasoning. For each instance, if the predicted and gold agreement labels do not match, the score is 0; if the agreement labels match, the instance receives a score of 1 only if both severity and edit type exactly match the gold labels, and 0 otherwise. The Composite Score is averaged across all instances and reflects end-to-end discrepancy reasoning performance. The evaluation code is available on our project’s GitHub repository\footnote{https://github.com/GeraldSun/RADAR}. 

We report results on (i) the \textbf{Mixed} set (natural edits combined with synthetic disagree edits introduced for evaluation balancing), which evaluates both clinical utility and safety since synthetic edits are constructed to be unsupported by the image evidence, and (ii) the \textbf{Natural} subset, which contains real radiologist revisions and reflects performance on real-world corrections without synthetic perturbations. We additionally report runtime, measured as the total time required for each experiment setting.

\section{Results}
Table \ref{tab:baseline_results} includes the performance results. On the Mixed dataset, Edit Type is consistently high across models (0.78–0.84), while Agreement (0.46–0.70) and Severity (0.46–0.56) show moderate variation. However, Composite Score remains comparatively low across settings (0.16–0.34), reflecting the difficulty of jointly satisfying agreement, severity, and edit type prediction within a single instance. For Gemini-3-Pro, 20 slices achieves the highest Agreement (0.693), while 50 slices yields the highest Composite Score (0.339). Similarly, Qwen3.5-plus does not exhibit monotonic gains from increasing slice counts; its best Composite Score occurs at 20 slices (0.255), with 50 slices slightly declining (0.245). Video-style input does not consistently improve results: it benefits Qwen3.5-plus but does not outperform slice-based settings for Gemini models. On the Natural dataset, Agreement must be interpreted cautiously due to strong class imbalance; for example, Gemini-2.5-Pro attains high Agreement likely influenced by the dominant “Agree” class. Gemini-3-Pro (50 slices) achieves the highest Composite Score (0.399). Overall, performance differences appear to be primarily driven by model family, and neither increasing slice count nor adopting video-style input reliably guarantees improved accuracy across settings. 

\begin{table}[!t]
\centering
\small
\caption{Baseline performance on RADAR. All values are accuracy for Agreement (Agr), Severity (Sev), Edit Type (Typ), and Composite Score (Comp). Comp requires exact match of Agr, Sev, and Typ.}
\setlength{\tabcolsep}{3pt}
\renewcommand{\arraystretch}{1.1}
\resizebox{\linewidth}{!}{
\begin{tabular}{lcccccccc c}
\hline
\multirow{2}{*}{\textbf{Model / Setting}} &
\multicolumn{4}{c}{\textbf{Mixed}} &
\multicolumn{4}{c}{\textbf{Natural}} &
\multirow{2}{*}{\textbf{Time (h)}} \\
\cline{2-9}
& \textbf{Agr} & \textbf{Sev} & \textbf{Typ} & \textbf{Comp}
& \textbf{Agr}$^{*}$ & \textbf{Sev} & \textbf{Typ} & \textbf{Comp}
& \\
\hline

\textbf{Gemini-2.5-Pro} \\
\quad 10 slices        & 0.522 & 0.467 & 0.814 & 0.193 & 0.791 & 0.412 & 0.856 & 0.301 & 1.1 \\
\quad 20 slices        & 0.489 & 0.464 & 0.807 & 0.168 & 0.784 & 0.425 & 0.837 & 0.288 & 1.2 \\
\quad 50 slices        & 0.460 & 0.526 & 0.814 & 0.248 & 0.680 & 0.562 & 0.850 & 0.386 & 1.3 \\
\quad video (all)      & 0.533 & 0.464 & 0.814 & 0.201 & 0.797 & 0.438 & 0.837 & 0.314 & 1.5 \\
\hline

\textbf{Gemini-3-Pro} \\
\quad 10 slices        & 0.686 & 0.536 & 0.803 & 0.321 & 0.647 & 0.614 & 0.830 & 0.386 & 1.8 \\
\quad 20 slices        & \textbf{0.693} & 0.544 & 0.781 & 0.321 & 0.660 & 0.614 & 0.791 & 0.379 & 2.0 \\
\quad 50 slices        & 0.675 & \textbf{0.560} & 0.821 & \textbf{0.339} & 0.627 & 0.614 & 0.843 & \textbf{0.399} & 3.0 \\
\quad video (all)      & 0.653 & 0.558 & 0.818 & 0.307 & 0.575 & \textbf{0.647} & 0.830 & 0.353 & 2.6 \\
\hline

\textbf{Qwen3.5-plus} \\
\quad 10 slices        & 0.558 & 0.533 & \textbf{0.839} & 0.223 & 0.418 & 0.601 & \textbf{0.869} & 0.216 & 3.3 \\
\quad 20 slices        & 0.591 & 0.536 & 0.818 & 0.255 & 0.477 & 0.621 & 0.824 & 0.255 & 3.4 \\
\quad 50 slices        & 0.613 & 0.518 & 0.818 & 0.245 & 0.510 & 0.569 & 0.830 & 0.242 & 3.7 \\
\quad video (all)      & 0.628 & 0.544 & \textbf{0.839} & 0.296 & 0.503 & 0.627 & 0.863 & 0.320 & 3.5 \\
\hline


\end{tabular}
}
\footnotesize{$^{*}$Agreement labels in the Natural subset are highly imbalanced across classes.}
\label{tab:baseline_results}
\end{table}

\section{Discussion} 

Our results highlight distinct challenges across the three components of discrepancy reasoning: agreement detection, severity assessment, and edit type classification. Edit type classification is comparatively easier, as it primarily relies on understanding the semantic function of an edit. In contrast, agreement detection requires robust cross-modal alignment to determine whether textual claims are supported by image evidence, particularly in the presence of synthetic edits designed to be unsupported by image evidence. Severity assessment remains the most challenging, as it demands calibrated clinical judgment to evaluate the practical impact of a discrepancy within real-world workflows. Our initial experiments suggest that while current multimodal models can effectively recognize linguistic patterns, reliable image-grounded and clinically nuanced reasoning remains difficult. A discrepancy-aware model could therefore serve as a verification layer, confirming image-supported corrections while flagging unsupported or clinically impactful edits before they influence downstream systems. Such capability may be particularly valuable in ED and resource-limited settings, where attending review may be delayed, and may ultimately strengthen the safety and robustness of AI-assisted clinical pipelines.

\section{Conclusion}
We present RADAR, a multimodal benchmark for image-grounded radiology report discrepancy analysis that evaluates agreement detection, severity assessment, and edit type classification within a realistic preliminary-to-attending review workflow. Our results show that while large multimodal models can effectively recognize linguistic edit patterns, reliable cross-modal verification and clinically calibrated severity reasoning remain challenging. Grouded in real preliminary-to-attending revisions, RADAR provides a clinically meaningful testbed for discrepancy-aware reasoning. While currently limited by its modest size and focus on a single modality and body site, RADAR establishes a foundation for broader multimodal discrepancy benchmarks. Future work will expand to additional modalities and anatomical regions and incorporate longitudinal cross-exam reasoning to further improve safety and workflow robustness.

\bibliographystyle{splncs04}
\bibliography{bibliography}

\end{document}